\documentclass[conference]{IEEEtran}
\IEEEoverridecommandlockouts
\usepackage{hyperref}
\usepackage{cite}
\usepackage{amsmath,amssymb,amsfonts}
\usepackage{algorithmic}
\usepackage{graphicx}
\usepackage{textcomp}
\usepackage{url}
\usepackage{xcolor}
\usepackage{booktabs}
\usepackage{caption}
\usepackage{mathrsfs}  
\def\BibTeX{{\rm B\kern-.05em{\sc i\kern-.025em b}\kern-.08em
    T\kern-.1667em\lower.7ex\hbox{E}\kern-.125emX}}
\begin{document}

\title{DAE-Fuse: An Adaptive Discriminative Autoencoder for
Multi-Modality Image Fusion
\thanks{* Equal Contribution, {\dag} Corresponding Author: wfsu@uic.edu.cn, {\ddag} Part of this work was done during Yuchen Guo’s internship at MMLab@SIAT, CAS.}
}

\author{
\IEEEauthorblockN{
Yuchen Guo\textsuperscript{*}\textsuperscript{\ddag} \quad 
Ruoxiang Xu\textsuperscript{*} \quad 
Rongcheng Li  \quad 
Weifeng Su\textsuperscript{\dag}
}
\IEEEauthorblockA{
\textsuperscript{}\textit{Department of Computer Science} \\
\textsuperscript{}\textit{Guangdong Provincial Key Lab of Interdisciplinary Research and Application for Data Science} \\
\textsuperscript{}\textit{Beijing Normal - Hong Kong Baptist University}
}
\noindent Project Page: \url{https://eurekaarrow.github.io/daefuse.github.io/}
}

\maketitle

\begin{abstract}
In extreme scenarios such as nighttime or low-visibility environments, achieving reliable perception is critical for applications like autonomous driving, robotics, and surveillance. Multi-modality image fusion, particularly integrating infrared imaging, offers a robust solution by combining complementary information from different modalities to enhance scene understanding and decision-making. However, current methods face significant limitations: GAN-based approaches often produce blurry images that lack fine-grained details, while AE-based methods may introduce bias toward specific modalities, leading to unnatural fusion results. To address these challenges, we propose DAE-Fuse, a novel two-phase discriminative autoencoder framework that generates sharp and natural fused images. Furthermore, We pioneer the extension of image fusion techniques from static images to the video domain while preserving temporal consistency across frames, thus advancing the perceptual capabilities required for autonomous navigation. Extensive experiments on public datasets demonstrate that DAE-Fuse achieves state-of-the-art performance on multiple benchmarks, with superior generalizability to tasks like medical image fusion.
\end{abstract}


\section{Introduction}
\label{sec:intro}
Multi-Modality Image Fusion (MMIF) is a key technique in the multimedia and low-level computer vision communities, aiming to integrate complementary information from different imaging modalities into a single, cohesive image. By leveraging the strengths of various modalities, MMIF enhances scene understanding and facilitates critical applications such as autonomous driving, robotics, and surveillance. Among these tasks, Infrared-Visible Image Fusion (IVIF) is particularly significant in extreme conditions, such as nighttime or low-visibility environments. Infrared images are effective for detecting thermal targets but lack detailed textures, while visible images capture rich textures but are sensitive to lighting. IVIF merges the strengths of both modalities to produce fused images that enhance environmental perception, offering more reliable information for decision-making in autonomous systems.

Over the years, MMIF methods have evolved significantly, with Generative Adversarial Networks (GANs) and Autoencoders (AEs) emerging as the predominant architectures for fusion tasks. In the context of the critical applications that we mentioned, Multi-Modality Object Detection (MMOD) is a great approach to evaluate the potential of the fusion results, which measure their ability to detect and classify objects, such as pedestrians, vehicles, and obstacles, under challenging conditions. MMOD plays a vital role in assessing the practical utility of fusion techniques, as accurate object detection directly impacts the safety and efficiency of autonomous systems. Fused images with superior performance in MMOD tasks enable autonomous vehicles to navigate complex environments with enhanced reliability, even under adverse scenarios like poor lighting or urban congestion.

GAN-based models leverage adversarial learning within a zero-sum game framework to fuse input images and source images. The typical strategy in MMIF tasks involves employing two discriminators: one to evaluate the fused image and the other to assess the source images. However, most methods either fuse two-dimensional image pairs directly before feeding them into the model or lack a well-designed feature extractor and corresponding loss function capable of effectively capturing distinct characteristics. This limitation weakens the feature extraction process. Consequently, these methods often produce fused images that are perceptually satisfactory—appearing distributionally similar to the original data—but fail to preserve fine-grained feature details, leading to blurriness within and between functional objects.

More efficient pipelines adopt an Autoencoder (AE)-based approach, integrating comprehensive feature extraction and reconstruction modules. These methods encode the two input modalities separately, combine the feature embeddings through channel concatenation, and decode the fused embeddings to generate the output image. With carefully crafted encoder blocks and loss functions, AE-based methods effectively extract both global and local features from different modalities. However, these methods typically share the same encoder and loss functions for both inputs and concatenate features directly, rather than combining them organically during the fusion phase. This can introduce modality bias, resulting in fused images that exhibit stronger traces from one modality while retaining less prominent information from the other. Experimental results confirm the presence of such bias, highlighting a key area for further improvement.


To address the aforementioned problems, we develop a more practical framework for image fusion tasks called the Discriminative Autoencoder (DAE), which enhances both feature extraction and the model’s overall perception of two modalities. We define the entire image fusion process as consisting of two phases.

In the Adversarial Feature Extraction Phase, the objective is to enhance the encoder’s ability to extract features by reconstructing single-modality images. We adopt a parallel Transformer-CNN architecture, which separately extracts shallow and deep features, and incorporate an adversarial game within the autoencoder. In this phase, two discriminative blocks are introduced to differentiate between the input images and their reconstructed counterparts, ensuring more effective feature extraction.

The second phase is the Attention-Guided Cross-Modality Fusion Phase, where a cross-attention module captures the complex correlations between the encoded images. The same discriminative blocks are utilized in this phase, but their role shifts to distinguishing between the source images and the fused image. Leveraging adversarial learning, the fused result evenly integrates information from both input modalities, avoiding bias toward any single modality. Together, these innovations form DAE-Fuse, a novel end-to-end discriminative autoencoder framework for multi-modality image fusion. By incorporating two-phase adversarial learning and a cross-attention fusion module, DAE-Fuse achieves a balanced fusion capability while demonstrating strong generalizability across diverse tasks and applications.

The main contributions are summarized as follows:

\begin{itemize}

\item {We propose a two-phase framework for general image fusion, comprising the adversarial feature extraction phase and the attention-guided cross-modality fusion phase, which enables robust high- and low-frequency feature extraction and produces natural fused images.}
\item {We design a novel fusion strategy that combines early fusion, implemented through a cross-attention module before reconstruction, with late fusion, achieved via adversarial learning between the discriminative blocks and the decoder.}
\item {Qualitative and quantitative experiments demonstrate that our model achieves state-of-the-art performance on both IVIF and MMOD public datasets. Furthermore, our approach generalizes effectively to Medical Image Fusion tasks.}

\end{itemize}

\section{Related Work}
Recently, the most widely used GAN \cite{goodfellow2020generative} based image fusion methods is to contrast two discriminators for fused images and both source images \cite{ddc,tarDAL}. For example, FusionGAN for IVIF task, in which the generator aims to generate fused images that major infrared intensities together with additional visible gradients and the discriminator aims to force the fused image to have more details existing just in the visible image \cite{fusiongan}. The adversarial game of it is designed between fused image and phase-contrast image \cite{GFPPC-GAN}. AE-based methods focuses on the elaborated reconstruction and fusion loss functions \cite{densefuse,Didfuse,Lrrnet}. For instance, \cite{densefuse} first adopted an AE-based model in the IVIF task. Moreover, \cite{Didfuse,CDD} proposed for high/low-frequency decomposition strategy aims to extract features more exactly for IVIF task in recent years.


\begin{figure}[htbp]
\centering
\includegraphics[width=\linewidth]{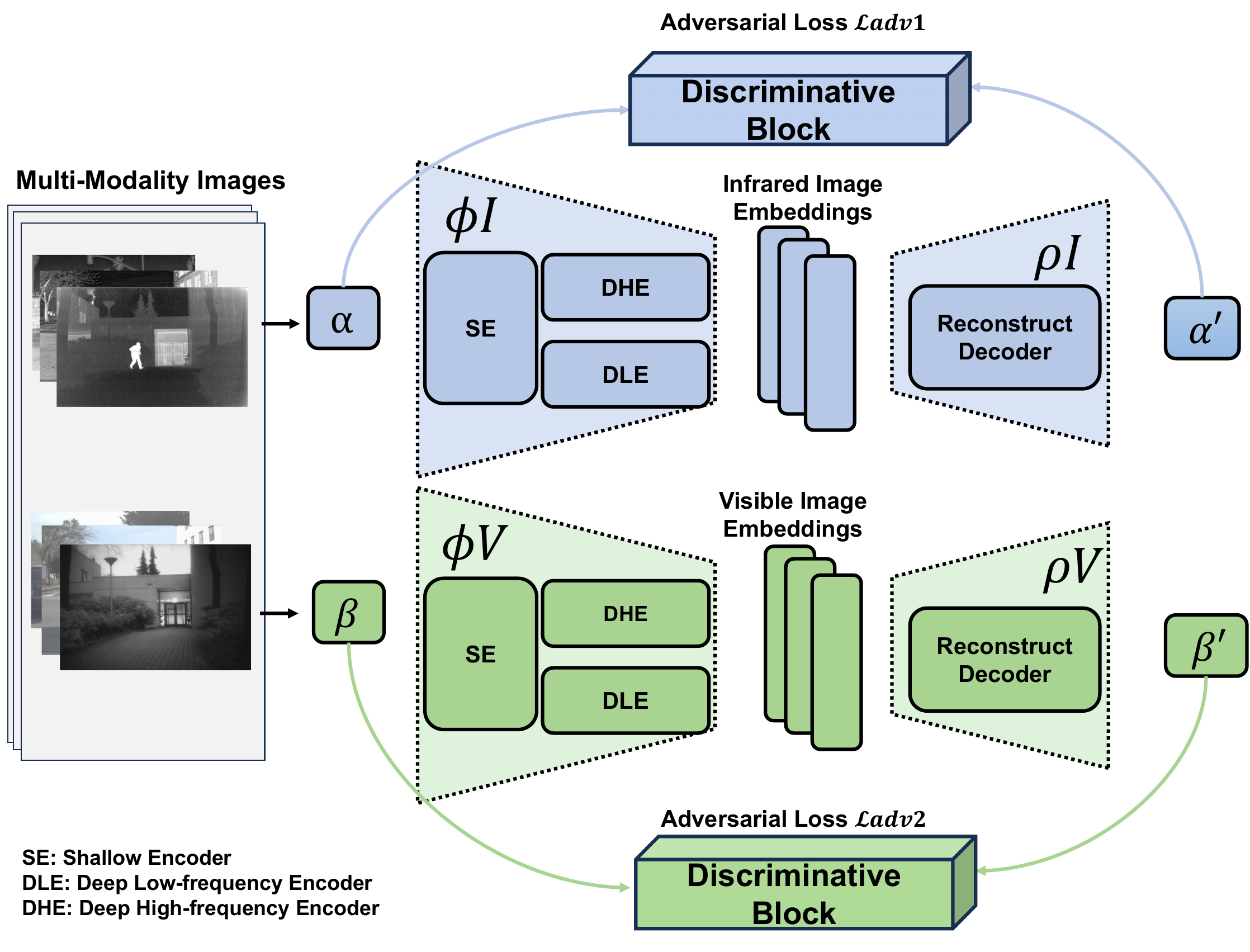} 
\caption{The workflow of the adversarial feature extraction phase. The cross-attention for fusion purpose is dismissed.}
\label{fig:stage1}
\end{figure}

\begin{figure}[htbp]
\centering
\includegraphics[width=\linewidth]{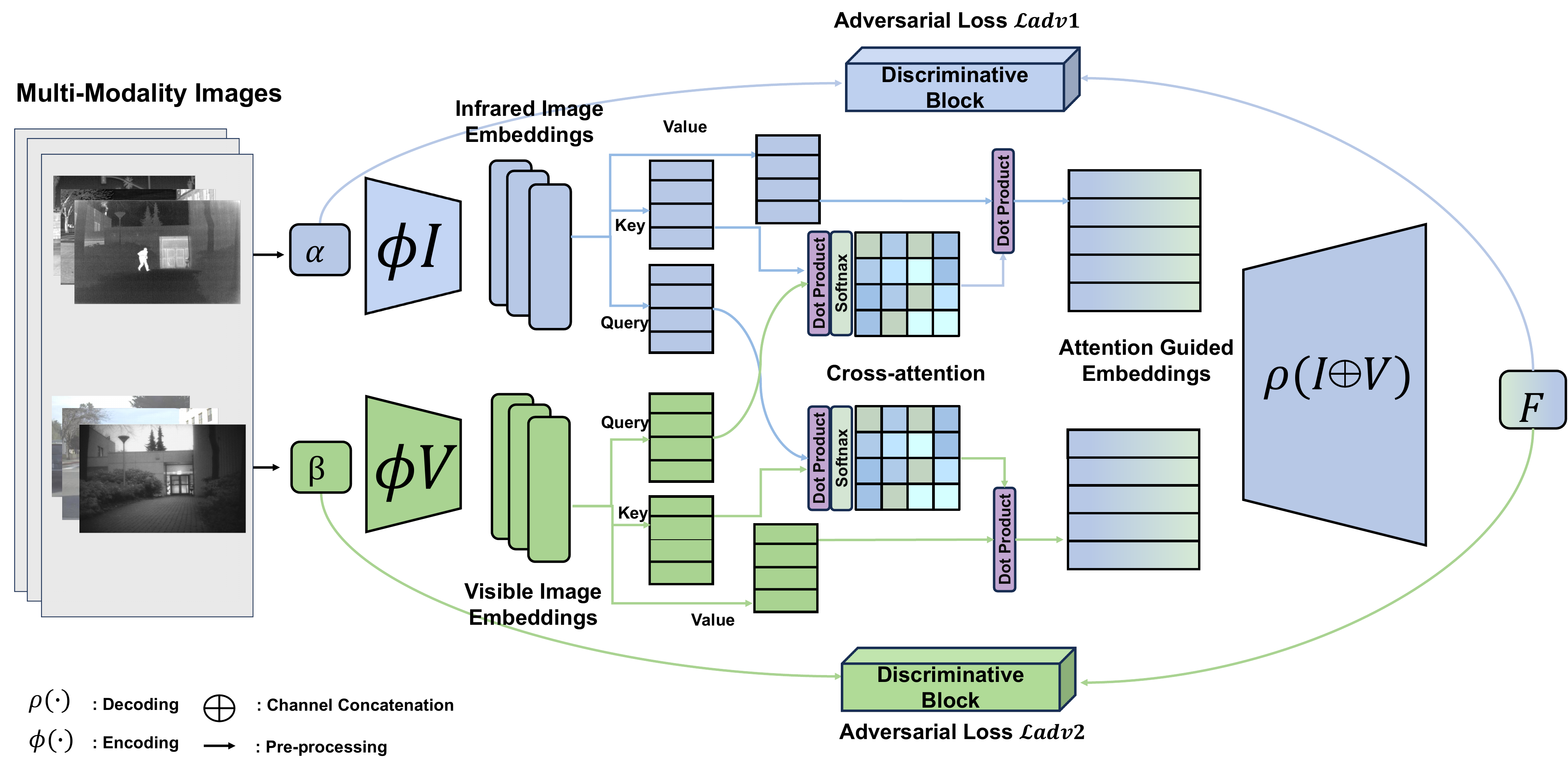} 
\caption{The workflow of the attention-guided cross-modality fusion phase.}
\label{fig:stage2}
\end{figure}

\section{Methods}

\subsection{Overview}
Due to the critical requirements for robustness and reliability in autonomous navigation systems, our DAE-Fuse is designed to effectively fuse visible and infrared images, thereby enhancing navigation safety, especially under challenging lighting conditions such as darkness, glare, or adverse weather. The architecture of DAE-Fuse is structured into two primary phases: the Adversarial Feature Extraction Phase (Phase 1) and the Attention-guided Cross-modality Fusion Phase (Phase 2). Our method processes video sequences frame by frame while ensuring temporal consistency, thus enabling effective video fusion for autonomous navigation applications.

\subsection{Adversarial Feature Extraction Phase}
To handle video sequences in autonomous navigation scenarios, we process consecutive frames while maintaining temporal consistency. This is achieved by implementing a temporal smoothing mechanism in the feature extraction process, ensuring stable fusion results across frames. Multi-level features are extracted by shallow and deep encoders. Specifically, to differentiate the various frequencies features, we deploy a Deep High-frequency Encoder (DHE, termed $\phi_{DH}(\cdot)$) and a Deep Low-frequency Encoder (DLE, termed $\phi_{DL}(\cdot)$) parallelly following the Shallow Encoder (SE, termed $\phi_S(\cdot,\cdot)$). Suppose the embedding from the encoding process is marked as: $\Phi(\cdot)$, and the input of first and second modalities as: $\alpha$, and $\beta$. The encoding process of paired $\{{\alpha, \beta}\}$ can be formulated as:
\begin{equation}
\begin{aligned}
    &\Phi(\alpha) = \mathit{C}[\phi_{DH}(\phi_{S}(\alpha)),  \phi_{DL}(\phi_{S}(\alpha))] \\
    &\Phi(\beta) = \mathit{C}[\phi_{DH}(\phi_{S}(\beta)),  \phi_{DL}(\phi_{S}(\beta))]
\end{aligned}
\end{equation}
where $\mathit{C}(\cdot, \cdot)$ donates channel concatenate operation.

Since the Transformer-based models are good at extracting low-frequency information while CNN-based models are sensitive to high-frequency information \cite{ICLR2022, NIPS2022}. We construct a Vision Transformer \cite{vit} for the DLE, and the DHE is implemented by a ResNet18 \cite{Resnet}. Restormer is a Channel-Transformer \cite{Restormer} architecture, which has achieved excellent performance in shallow region reconstruction task without increasing too much computation, so we use a channel-Transformer block for SE to extract shallow features. While the Reconstruction Decoder (RD, termed $\rho_R(\cdot)$) is responsible for reconstructing the embeddings to image. RD shares the same architecture with SE. The decoding process of paired $\{{\alpha, \beta}\}$ can be formulated as:
\begin{equation}
    \tilde{\alpha} = \rho_R(\alpha), \tilde{\beta} = \rho_R(\beta)
\end{equation}
where $\tilde{\alpha}$ and $ \tilde{\beta}$ represent the images $\alpha$ and $\beta$ after reconstructing, respectively.

The adversarial process is implemented by two discriminative blocks from different modalities (DM1 and DM2, termed $D_{M1}(\cdot ,\cdot)$ and $D_{M2}(\cdot ,\cdot)$ respectively). Discriminative blocks is implemented by a stack of Con2D-LeackyReLU-BatchNorm layers and a fully connected layer.
Accordingly, the adversarial learning process can be formulated as minimizing the following adversarial objective:
\begin{equation}
\begin{aligned}
\min_{AE} \max_{D_{M1}, D_{M2}} \Big( & \mathbb{E}[\log (D_{M1}(\alpha))] + \mathbb{E}[\log (D_{M2}(\beta))] & \\
 + \mathbb{E}[ \log &  (1- (D_{M1}(\tilde{\alpha}))]  + \mathbb{E}[ \log (1- (D_{M2}(\tilde{\beta}))]  \Big)  &
\end{aligned}
\end{equation}

\subsection{Attention-guided Cross-modality Fusion Phase}
In this phase, we developed a feature aggregation strategy by calculating the cross-attention weights \cite{vit}. We use the same structure of discriminative blocks as before. During the adversarial fusion step, the inputs of a discriminative block are a fused image two source images. In autonomous navigation scenarios, different modalities often capture complementary information - infrared images excel at detecting pedestrians and objects in low-light conditions, while visible images provide rich texture and context information in normal lighting conditions. Our cross-modality attention module is specifically designed to leverage these complementary strengths, enabling robust object detection across various lighting conditions.

\subsubsection{Early Fusion} 
Owing to the data gap between different modalities, current approaches in MMIF are limited to only incorporating element-wise additions for extracted feature embeddings, which does not capture the important interactions. We deploy a cross-modality attention module, making the different embeddinga can naturally interact another modality before fusion.
After extracting features from encoders ($\Phi(\alpha)$, $\Phi(\beta)$), embeddings of images from two modalities are obtained. Here we use the embedding of  $\alpha$ as the Query $Q$, while the embeddings of $\beta$ as the Key $K$ and the Value $V$. Assuming the attention guided embeddigns are denoted as: $\hat{(\Phi(\alpha))}$ and $\hat{(\Phi(\beta))}$.
The mapping can be expressed as:
\begin{flalign}
&\text{CroAttn}_{\Phi(\alpha) \rightarrow \Phi(\beta)}(\Phi(\alpha), \Phi(\beta)) = \text{softmax}\left(\frac{QK^T}{\sqrt{d_k}}\right) & \nonumber \\
&= \text{softmax}\left(\frac{(\text{\textbf{W}}_q \Phi(\alpha) \Phi(\beta)^T \text{\textbf{W}}_k^T)}{\sqrt{d_k}}\right) \text{\textbf{W}}_v\Phi(\beta) & \nonumber \\
&\rightarrow A_{\text{croattn}} \text{\textbf{W}}_v(\Phi(\beta))\rightarrow \hat{(\Phi(\beta))} &
\end{flalign}

where $\text{\textbf{W}}_q$, $\text{\textbf{W}}_v$ and $\text{\textbf{W}}_k$ are trainable weight matrices multiplied to queries ($Q_{\Phi(\alpha)}$) and key-value pair ($K_{\Phi(\beta)}$, $V_{\Phi(\beta)}$), and $A_{\text{croattn}}$ is the cross-attention matrix for computing the weighted average of ${\Phi(\beta)}$.
\subsubsection{Adversarial Fusion}
First, the decoder generates fused image from attention-guided embeddings:
\begin{equation}
    \mathscr{F}(\alpha , \beta) = \rho_R[\mathit{C}(\hat{\Phi(\alpha)}, \hat{\Phi(\beta)})]   
\end{equation}

Then, the adversarial process is adapted to following formulation:
\begin{equation}
\begin{aligned}
\min_{AE} \max_{D_{M1}, D_{M2}} \Big( & \mathbb{E}[\log (D_{M1}(\alpha))] + \mathbb{E}[\log (D_{M2}(\beta))] & \\
 + \mathbb{E}[ \log  (1- (D_{M1}( &\mathscr{F}  (\alpha , \beta)))]  + \mathbb{E}[ \log (1- (D_{M2}(\mathscr{F}(\alpha , \beta)))]  \Big)  &
\end{aligned}
\end{equation}

\subsection{Loss Function}

\subsubsection{Temporal Consistency Loss}
To ensure smooth transitions between consecutive frames in video sequences, we introduce an temporal consistency loss for both phase one and phase two:
\begin{equation}
\mathcal{L}{{temp}} = \frac{1}{HW} \left\lVert \mathscr{F}(\alpha_t, \beta_t) - \mathscr{F}(\alpha_{t-1}, \beta_{t-1}) \right\rVert_1
\end{equation}
where $\mathscr{F}(\alpha_t, \beta_t)$ represents the fused result at current frame $t$, and $\mathscr{F}(\alpha_{t-1}, \beta_{t-1})$ represents the fused result from the previous frame.

\subsubsection{Phase one}
We construct the loss function for the autoencoder and discriminative blocks separately. The loss function of AE is divided into two parts: adversarial loss and content loss: 
\begin{equation}
\mathcal{L}{_{AE}}{^I} = \lambda\mathcal{L}{_{AE}}{^{advI}} + \sigma\mathcal{L}_{Enc}^{correlation} + (1-\sigma)\mathcal{L}{_{Dec}}^{content}
\end{equation}
where the $\sigma$ is the hyper-parameter.
And the adversarial loss for encoder-decoder is:
\begin{equation}
\mathcal{L}{_{AE}}{^{advI}} = \mathbb{E}[ \log (1- (D_{M1}(\tilde{\alpha}))]  + \mathbb{E}[ \log (1- (D_{M2}(\tilde{\beta}))]
\end{equation}
Additionally, we use correlation decomposition loss \cite{CDD} for differentiate high-frequency feature and low-frequency feature:
\begin{equation}
\mathcal{L}_{Enc}^{correlation} = \frac{(CC (\phi_{DH}(\alpha), \phi_{DH}(\beta)^2}{CC (\phi_{DL}(\alpha), \phi_{DL}(\beta)) + \epsilon}
\end{equation}
where $\epsilon$ set 1.01 to ensure the result always be positive.

The decoder reconstruction loss function consists of the square of the L2 norm and structural similarity index:
\begin{equation}
\mathcal{L}_{Dec}{^{content}} = \lVert \alpha - D(\alpha) \rVert_2^2 + (1-SSIM(\alpha, D(\alpha)))
\end{equation}
The adversarial loss of discriminative block DM1 and DM2 are of same structure. Take DM1 as an example:
\begin{equation}
\begin{aligned}
 \mathcal{L}_{DM1}{^{advI}} =  \mathbb{E}[-\log (D_{M1}(\alpha))]  + \mathbb{E}[ -\log (1- (D_{M1}(\tilde{\alpha}))] 
\end{aligned}
\end{equation}

\subsubsection{Phase two}
Since the inputs of discriminative blocks have been changed, here we represent the adversarial loss of phase two. Also, we use the structural content loss \cite{tang2022image} as:
\begin{equation}
\mathcal{L}{_{text}}{^{II}}  = \frac{1}{HW} \left\lVert \nabla I_f - \max(\lvert \nabla \alpha \rvert, \lvert \nabla \beta \rvert) \right\rVert_1
\end{equation}
and ,
\begin{equation}
\mathcal{L}{_{int}}{^{II}} = \frac{1}{HW} \left\lVert I_f - \max(\alpha, \beta) \right\rVert_1
\end{equation}
Thus, the whole losses in phase two can formulated as:
\begin{equation}
\mathcal{L}{^{II}} = \mathcal{L}{_{DM}}{^{advII}} + \mathcal{L}{_{AE}{^{II}}}
\end{equation}

\begin{figure}[htbp]
\centering
\includegraphics[width=\linewidth]{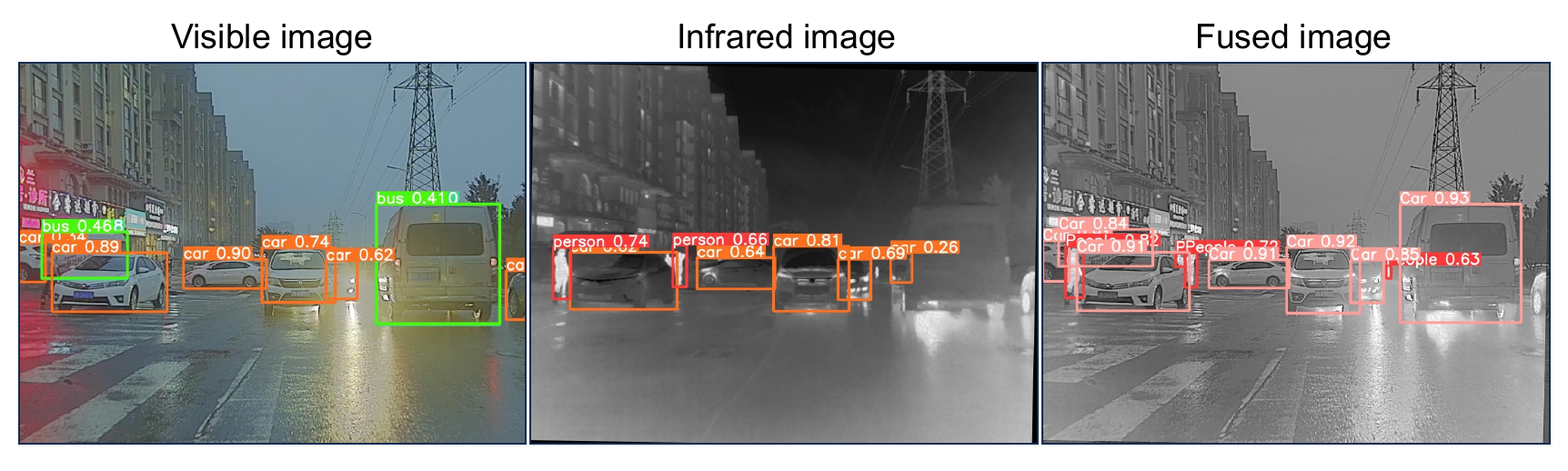} 
\caption{Object detection ability of DAE-Fuse: the visible image can detect the car in the right but fail to capture the people; the infrared displays an opposite ability on this two objects; and the fused image from DAE-Fuse successfully detects all of them.}
\label{fig:mmod_qua}
\end{figure}

\begin{figure}[htbp]
\centering
\includegraphics[width=\linewidth]{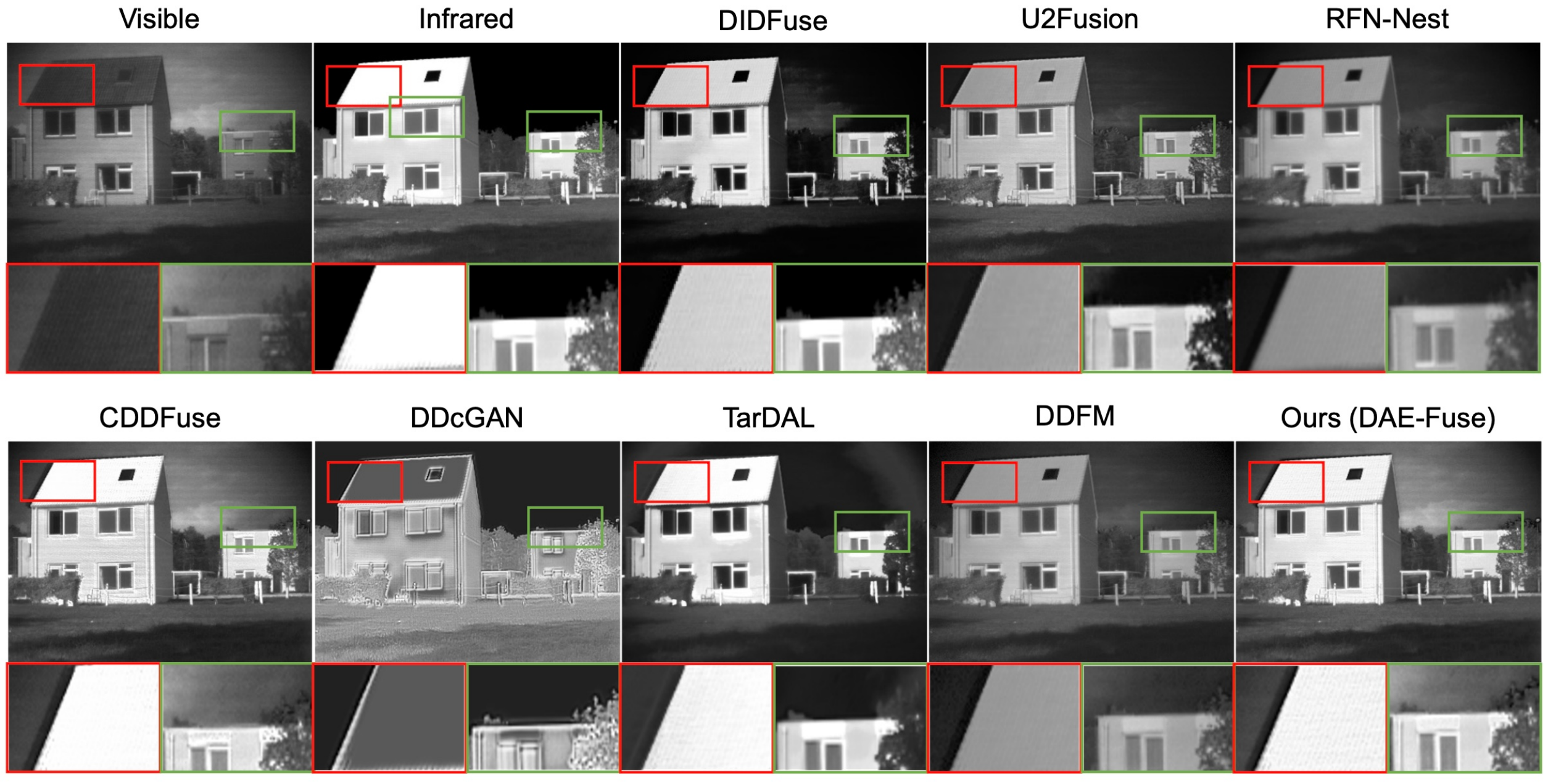} 
\caption{Qualitative comparison with state-of-the-art methods on TNO dataset.}
\label{fig:quali_ivif}
\end{figure}

\begin{figure}[t]
\centering
\includegraphics[width=\linewidth]{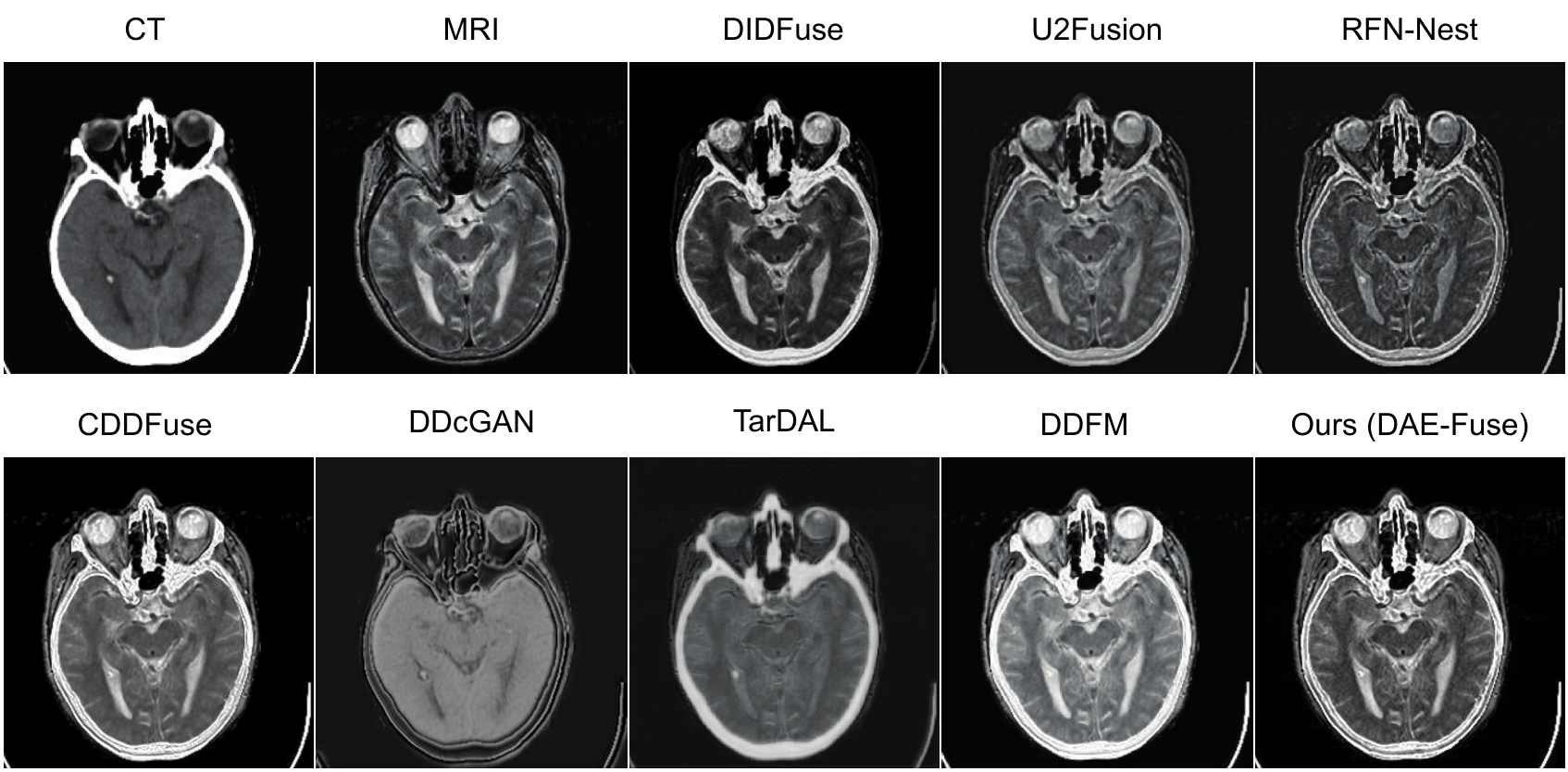} 
\caption{Qualitative comparison with state-of-the-art methods on MRI-CT dataset.}
\label{fig:quali_mif}
\end{figure}

\section{Experiments}

\subsection{Setup}
\subsubsection{Datasets and metrics} The IVIF experiments contain three popular benchmarks to measure the fusion model performance, including MSRS \cite{30}, RoadScene \cite{31}, and TNO \cite{32}. There are usually eight unsupervised metrics to measure the quantitative performance including entropy (EN) \cite{ddc_50}, standard deviation (SD) \cite{ddc_51}, spatial frequency (SF) \cite{cdd_40}, visual information fidelity (VIF) \cite{ddc_53}, sum of correlation of differences (SCD) \cite{cdd_40}, mutual information (MI) \cite{cdd_40}, $Q^{AB/F}$ \cite{cdd_40} and structural similarity index measure (SSIM) \cite{ddc_52}. For the MMOD task, we use the mAP@50(\%) metric to measure our detection result. Higher metrics value indicates a better result. Our experiments were implemented based on the PyTorch framework and performed on a server with an NVIDIA A100 GPU.

\subsubsection{Implementation Details.}  In the first phase, we trained 80 epochs for both the autoencoder and discriminator, while in the second phase, the model was trained with 140 epochs. We use the Adam optimizer for the autoencoder and the RMSProp optimizer for the discriminator with an initial learning rate of 1e-4 and decreasing by 0.5 every 20 epochs. The training samples are randomly cropped into 128 × 128 patches in preprocessing stage and the batch size is set to 16. In the MMOD downstream testing, the generated 4200 fused images is partitioned into training, validation, and test sets, with an 8:1:1 ratio for a YOLOv8n \cite{29}.

\subsection{Infrared-Visible Image Fusion}
We tested our model on the three IVIF datasets and compared them with seven state-of-the-art methods including DIDFuse \cite{Didfuse}, U2Fusion \cite{U2F}, RFN-Nest \cite{RFN-nest}, DDcGAN \cite{ddc}, TarDAL \cite{tarDAL}, CDDFuse \cite{CDD} and DDFM \cite{DDFM}.

\subsection{Infrared-Visible Video Fusion}
Video fusion is tested on the RoadScene dataset, which you can find the sample result from the supplemental material.

\begin{table}[h]
  \centering
  \captionsetup{skip=0pt}
  \caption{Quantitative comparisons on TNO, MSRS, and RoadScene datasets. Bold \textcolor{red}{red} indicates the best, and bold \textcolor{blue}{blue} indicates the second best.}
  \label{tab:all-IVIF-datasets}

    \centering
    \textbf{Dataset: TNO}
    \label{tab:TNO-Transposed}
    \begin{tabular}{llllllll}
        \toprule
         & EN & SD & SF & MI & SCD & VIF & Qabf \\ 
        \midrule
        DIDFuse & 6.97 & 45.12 & 12.59 & 1.70 & 1.71 & 0.60 & 0.40 \\
        U2Fusion & 6.83 & 34.55 & 11.52 & 1.37 & 1.71 & 0.58 & 0.44 \\
        RFN-Nest & 6.83 & 34.50 & \textcolor{red}{15.71} & 1.20 & 1.67 & 0.51 & 0.39 \\
        DDcGAN & 6.78 & 46.33 & 9.12 & 1.78 & 1.72 & 0.48 & 0.35 \\
        TarDAL & 6.84 & 45.63 & 8.68 & \textcolor{blue}{1.86} & 1.52 & 0.53 & 0.32 \\
        CDDFuse & \textcolor{blue}{7.12} & 46.00 & 13.15 & 2.19 & \textcolor{blue}{1.76} & 0.77 & \textcolor{blue}{0.54} \\
        DDFM & 7.06 & \textcolor{red}{51.42} & 13.03 & 2.21 & 1.66 & \textcolor{red}{0.81} & 0.49 \\
        \textbf{Ours} & \textcolor{red}{7.17} & \textcolor{blue}{46.63} & \textcolor{blue}{13.31} & \textcolor{red}{2.23} & \textcolor{red}{1.89} & \textcolor{blue}{0.79} & \textcolor{red}{0.57} \\
        \bottomrule
    \end{tabular}

  \vspace{1em} 

    \centering
    \textbf{Dataset: MSRS}
    \label{tab:MSRS-Transposed}
        \begin{tabular}{llllllll}
        \toprule
         & EN & SD & SF & MI & SCD & VIF & Qabf \\ 
        \midrule
        DIDFuse & 4.27 & 31.49 & 10.15 & 1.61 & 1.11 & 0.31 & 0.20 \\
        U2Fusion & 5.37 & 25.52 & 9.07 & 1.40 & 1.24 & 0.54 & 0.42 \\
        RFN-Net & 5.56 & 24.09 & \textcolor{blue}{11.98} & 1.30 & 1.13 & 0.51 & 0.43 \\
        DDcGAN & 6.54 & 41.33 & 7.68 & 2.36 & 1.47 & 0.48 & 0.57 \\
        TarDAL & 5.28 & 25.22 & 5.98 & 1.49 & 0.71 & 0.42 & 0.18 \\
        CDDFuse & 6.70 & \textcolor{blue}{43.38} & 11.56 & \textcolor{blue}{3.47} & \textcolor{blue}{1.62} & \textcolor{red}{1.05} & \textcolor{blue}{0.69} \\
        DDFM & \textcolor{red}{6.88} & 40.75 & 11.65 & 2.35 & 1.62 & 0.81 & 0.58 \\
        \textbf{Ours} & \textcolor{blue}{6.86} & \textcolor{red}{44.05} & \textcolor{red}{12.26} & \textcolor{red}{3.48} & \textcolor{red}{1.75} & \textcolor{blue}{0.98} & \textcolor{red}{0.72} \\
        \bottomrule
        \end{tabular}
  
    \vspace{1em} 

    \centering
    \textbf{Dataset: RoadScene}
    \label{tab:RoadScene-Transposed}
        \begin{tabular}{llllllll}
        \toprule
         & EN & SD & SF & MI & SCD & VIF & Qabf \\ 
        \midrule
        DIDFuse & 7.43 & 51.58 & 14.66 & 2.11 & 1.70 & 0.58 & 0.48 \\
        U2Fusion & 7.09 & 38.12 & 13.25 & 1.87 & 1.70 & 0.60 & 0.51 \\
        RFN-Nest & 7.21 & 41.25 & 16.19 & 1.68 & 1.73 & 0.54 & 0.45 \\
        DDcGAN & 7.20 & 38.32 & 11.68 & 2.25 & 1.56 & 0.48 & 0.31 \\
        TarDAL & 7.17 & 47.44 & 10.83 & 2.14 & 1.55 & 0.54 & 0.40 \\
        CDDFuse & \textcolor{blue}{7.44} & \textcolor{blue}{54.67} & \textcolor{blue}{16.36} & 2.30 & \textcolor{blue}{1.81} & 0.69 & 0.52 \\
        DDFM & 7.41 & 52.61 & 13.57 & \textcolor{red}{2.35} & 1.66 & \textcolor{blue}{0.75} & \textcolor{blue}{0.65} \\
        \textbf{Ours} & \textcolor{red}{7.57} & \textcolor{red}{56.05} & \textcolor{red}{17.12} & \textcolor{blue}{2.32} & \textcolor{red}{1.85} & \textcolor{red}{0.76} & \textcolor{red}{0.68} \\
        \bottomrule
        \end{tabular}
  
\end{table}
\subsubsection{Qualitative comparisons}
As shown in Figure \ref{fig:quali_ivif}, we select a representative scenario to evaluate feature extraction capabilities and highlight model biases. The GAN-based models DDcGAN and TarDAL exhibit a blurry style that loses many details, while the AE-based method DIDFuse shows a clear bias toward infrared images, rendering the sky and grass excessively dark. In contrast, DAE-Fuse achieves the best outcome, preserving rich texture details and maintaining a seamless balance between modalities. Notably, it fully retains the roof’s texture, a result unmatched by other methods. Although CDDFuse, with its parallel encoder architecture, maintains some roof details, parts are overexposed, and additional noise is introduced to the wall textures. By comparison, DAE-Fuse naturally fuses the walls from both inputs, demonstrating superior fusion quality.

\subsubsection{Quantitative comparisons}
Afterward, we use the seven metrices to quantitively compare the results with other models, which are displayed in Table \ref{tab:all-IVIF-datasets}. DAE-Fuse shows an outstanding performance across all the measurement indices, demonstrating the effectiveness of our method.

\begin{table}[t]
    \centering
    \captionsetup{skip=1pt}
    \caption{Quantitative comparisons on MRI-CT datasets. Bold \textcolor{red}{red} indicates the best, and bold 
    \textcolor{blue}{blue} indicates the second best.}
    \label{tab:all-MIF-datasets}
    
    \textbf{Dataset: MRI-CT} \\
    \begin{tabular}{llllllll}
    \toprule
        ~ & EN & SD & SF & MI & SCD & VIF & Qabf \\ 
    \midrule
        DIDFuse & 4.37 & 58.34 & \textcolor{blue}{34.64} & 1.71 & 0.69 & 0.41 & 0.38 \\ 
        U2Fusion & 4.21 & 61.98 & 32.54 & 2.08 & 0.75 & 0.37 & 0.46 \\ 
        RFN-Nest & \textcolor{red}{4.97} & 70.36 & 33.42 & 1.98 & 0.68 & 0.43 & 0.52 \\ 
        DDcGAN & 4.26 & 62.56 & 30.61 & 1.72 & 0.65 & 0.38 & 0.42 \\ 
        TarDAL & 4.35 & 61.14 & 28.38 & 1.94 & 0.92 & 0.32 & \textcolor{blue}{0.56} \\ 
        CDDFuse & 4.49 & \textcolor{blue}{71.36} & 34.02 & \textcolor{blue}{2.16} & \textcolor{blue}{1.18} & \textcolor{blue}{0.44} & \textcolor{blue}{0.56} \\ 
        DDFM & 4.77 & 69.35 & 32.77 & 1.98 & 1.03 & 0.41 & 0.54 \\ 
        \textbf{Ours} & \textcolor{blue}{4.83} & \textcolor{red}{76.19} & \textcolor{red}{35.56} & \textcolor{red}{2.20} & \textcolor{red}{1.21} & \textcolor{red}{0.49} & \textcolor{red}{0.57} \\ 
    \bottomrule
    \end{tabular}
    \vspace{1em}

\end{table}

{\small
\begin{table}
  \caption{Quantitative comparison of our method with 7 state-of-the-art methods in AP@50(\%) values for MM object detection on M\textsuperscript{3}FD dataset. Bold \textcolor{red}{red} indicates the best, and Bold \textcolor{blue}{blue} indicates the second best.}
  \label{tab:mmod_qua}
    \begin{tabular}{lccccccccc} 
    \toprule
    & Peo & Car & Lam & Bus & Mot & Tru & mAP \\
    \midrule
    Ir & 0.804 & 0.886 & 0.712 & 0.802 & 0.725 & 0.743 & 0.779 \\
    Vis & 0.721 & 0.865 & 0.848 & 0.811 & 0.794 & 0.791 & 0.805 \\
    DID & 0.791 & 0.924 & 0.857 & 0.833 & 0.787 & 0.788 & 0.830 \\
    U2F & 0.802 & 0.922 & \textcolor{blue}{0.870} & 0.839 & 0.783 & 0.786 & 0.833 \\
    RFN & 0.813 & 0.915 & 0.851 & 0.829 & \textcolor{blue}{0.813} & 0.875 & 0.849 \\
    DDc & 0.797 & 0.908 & 0.832 & 0.895 & 0.805 & 0.872 & 0.851 \\
    TarD & 0.835 & \textcolor{red}{0.947} & 0.854 & 0.928 & 0.811 & 0.874 & 0.874 \\
    CDD & \textcolor{blue}{0.846} & 0.928 & 0.864 & \textcolor{blue}{0.931} & \textcolor{blue}{0.813} & \textcolor{red}{0.891} & \textcolor{blue}{0.878} \\
    DDFM & 0.837 & 0.926 & 0.869 & 0.927 & 0.809 & 0.882 & 0.875 \\
    \textbf{Ours} & \textcolor{red}{0.855} & \textcolor{blue}{0.931} & \textcolor{red}{0.874} & \textcolor{red}{0.949} & \textcolor{red}{0.822} & \textcolor{blue}{0.890} & \textcolor{red}{0.887} \\
    \bottomrule
    \end{tabular}
\end{table}

\subsection{Object Detection Ability}
A single modality image \emph{i.e.}, an individual infrared image usually lacks certain features of objects during the detection process. As shown in Figure \ref{fig:mmod_qua}., infrared images exhibit a robust capability for detecting humans but may overlook objects that do not produce thermal radiation. On the other hand, visible images struggle to recognize humans due to reflective lights from vehicles and lamps. After fusing the images from two modalities, by combining the advantages of two types of features, both humans and vehicles are well detected in the fusion image. Table \ref{tab:mmod_qua} exhibits the detection results. Benefiting from the excellent fusion ability on the thermal radiation information, DAE-Fuse successfully detected all the targets and outperformed all other SOTA methods.

\subsection{Generalization on Medical Image Fusion tasks}
To validate the generalizability and robustness of our DAE-Fuse, we used the same models in our IVIF testing to fuse medical images.



\subsubsection{Performance}
The qualitative and quantitative comparisons for source images and fusion results on the MRI-CT dataset are shown in Figure \ref{fig:quali_mif} and Table \ref{tab:all-MIF-datasets}, respectively. Similar to the IVIF experiments, GAN-based models produce blurry results with significant detail loss, while AE-based methods generally perform better in feature extraction. For example, U2Fusion and RFN-Nest assign lower weight to CT images, leading to darker target outlines, while DIDFuse emphasizes CT features more conspicuously. In contrast, DAE-Fuse achieves superior integration of information from both modalities, preserving critical details and maintaining a balanced fusion. Quantitatively, DAE-Fuse outperforms other models on nearly all metrics, demonstrating its generalizability and adaptability to MIF tasks without requiring adjustments.

\subsection{Ablation study}
The ablation study is performed on the TNO dataset. We use EN, SD, VIF and Qabf to present the quantitative result, which is illustrated in Table \ref{tab:ablation}.

\subsubsection{Discriminative blocks}
In this experiment, we removed the discriminator blocks while keeping all other configurations unchanged. In phase one, performance decreased across all metrics, highlighting the role of discriminative blocks in guiding feature extraction and improving image quality. Similarly, in phase two, their contribution to the final outcomes demonstrates the importance of the adversarial game between the fused results and input pairs.

\subsubsection{Cross-attention module}
To verify the contribution of the cross-attention module, we conducted an experiment by abandoning it and using only the channel concatenation in phase two with the discriminative blocks reserved. The result shows that all the matrices decrease. The interaction between the embeddings of two modalities brought by the cross-attention did boost the fusion results.

{\small
\begin{table}[h]
\centering
  \caption{Ablation comparisons of our method on TNO dataset. Bold \textcolor{red}{red} indicates the best.}
\begin{tabular}{lcccc}
\toprule
Configurations & EN & SD & VIF & Qabf \\
\midrule
w/o discriminative blocks in phase one & 6.81 & 43.20 & 0.91 & 0.63 \\
w/o discriminative blocks in phase two & 6.78 & \textcolor{red}{45.31} & 0.83 & 0.61 \\
w/o cross-attention module & 6.78 & 43.79 & 0.89 & 0.62 \\
\textbf{Ours} & \textcolor{red}{6.86} & 44.05 & \textcolor{red}{0.98} & \textcolor{red}{0.68} \\
\bottomrule
\label{tab:ablation}
\end{tabular}
\end{table}
}

\section{Conclusion}
In conclusion, our novel DAE-Fuse framework, tailored for autonomous navigation, effectively overcomes the limitations of traditional image fusion methods by generating sharp and natural fused images. This is accomplished through a two-phase approach: adversarial feature extraction and attention-guided cross-modality fusion. The integration of discriminative blocks and a cross-attention module within our encoder-decoder architecture ensures enhanced feature extraction and maintains structural integrity, resulting in visually appealing fused images. Evaluations on diverse datasets confirm DAE-Fuse's superiority, not only in standard metrics but also in improving object detection for autonomous vehicles, setting a new standard in image fusion for real-world applications.

\section{Acknowledgement}
Our work was supported in part by the Guangdong Provincial Key Laboratory of IRADS (2022B1212010006) and in part by Guangdong Higher Education Upgrading Plan (2021-2025) with No.of UICR0400006-24, UICR0400005-24.

\bibliographystyle{IEEEbib}
\bibliography{icme2025references}

\vspace{12pt}

\end{document}